%% file: main.tex
\documentclass[10pt,twocolumn,letterpaper]{article}

\usepackage[pagenumbers]{wacv}

\input{preamble}

\definecolor{wacvblue}{rgb}{0.21,0.49,0.74}
\usepackage[pagebackref,breaklinks,colorlinks,allcolors=wacvblue]{hyperref}
\hypersetup{
  pdftitle={DeltaS: Reading the Gated Linear Attention State for KV Cache Eviction in Streaming Video},
  pdfauthor={Taeyoun Kwon, Seungjin Kim, Hyeonyu Kim, Moon Hwan Kim}
}

\title{DeltaS: Reading the Gated Linear Attention State\\for KV Cache Eviction in Streaming Video}

\author{
Taeyoun Kwon$^{1,2}$\quad
Seungjin Kim$^{1,2}$\quad
Hyeonyu Kim$^{1}$\quad
Moon Hwan Kim$^{1,3}$\\[4pt]
$^1$Maum AI Inc.\quad
$^2$Seoul National University\quad
$^3$Yonsei University\\[2pt]
{\tt\small \{taeyoun.kwon, seungjin.kim, hykim, mhk\}@maum.ai}
}

\begin{document}
\maketitle

\input{sec/0_abstract}
\input{sec/1_intro}
\input{sec/2_related}
\input{sec/3_method}

\input{sec/4_experiments}
\input{sec/5_conclusion}

\FloatBarrier

{
    \small
    \bibliographystyle{ieeenat_fullname}
    \bibliography{main}
}

\clearpage
\appendix
\input{sec/A_ports}
\input{sec/B_streaming}

\end{document}

%% file: preamble.tex
\usepackage{makecell}
\usepackage{colortbl}
\usepackage{algorithm}
\usepackage{algpseudocode}
\algrenewcommand\algorithmicrequire{\textbf{Input:}}
\usepackage{enumitem}
\usepackage{placeins}

\input{tables/tabledefs}

%% file: tables/tabledefs.tex
\definecolor{oursbg}{gray}{0.90}
\newcommand{\oc}{\cellcolor{oursbg}}
\newcommand{\ocL}[1]{\multicolumn{1}{>{\columncolor{oursbg}[4pt][4pt]}l}{#1}}
\newcommand{\m}[1]{\hspace{0.85em}#1}
\newcommand{\band}[1]{\multicolumn{4}{l}{\textbf{#1}}\\[1pt]}
\newcommand{\hd}[1]{\makebox[48pt]{\makecell{#1}}}
\newcommand{\hda}[1]{\makebox[48pt]{#1}}
\newcommand{\hdm}[1]{\makebox[73pt][l]{#1}}
\newcommand{\hscore}[1]{\makebox[64pt][l]{#1}}
\newsavebox{\mtbox}

%% file: sec/0_abstract.tex
\begin{abstract}
Recent video-language models increasingly adopt hybrid architectures that interleave linear and full attention layers for efficient long-context processing. While the recurrent state of linear attention remains fixed in size, the KV cache of full attention continues to grow with the video stream, making eviction necessary under a bounded memory budget. The key challenge in streaming is that eviction must occur before the question arrives, so what to retain has to be decided without the question. Existing eviction methods derive token scores from the KV cache itself, using position, attention, or key-value representations, and attention-based scores further require proxy queries or extra computation. Hybrid backbones offer another source of signal. In gated-delta linear attention, the recurrent state is updated by the residual between each input and what can already be retrieved from the state, so its change over a chunk of frames reflects how much new information the chunk brings. We propose DeltaS, a query-agnostic, training-free method that retains video chunks inducing larger normalized state change, or state drift. In a controlled comparison with the budget and retention policy held fixed, state drift outperforms position-, attention-, and key-value-based signals. With a signal costing only 1.9\% of the forward pass, DeltaS surpasses the strongest query-agnostic bounded-memory baseline by 2.1 points on average across six long-video benchmarks and by 5.6 points on the longest benchmark. These results suggest that the two memories of hybrid architectures can work cooperatively. Code is available at {\def\UrlBreaks{\do\/}\def\UrlBigBreaks{}\url{https://github.com/MaumAI-Company/DeltaS}}.
\end{abstract}

%% file: sec/1_intro.tex
\section{Introduction}
\label{sec:intro}

\begin{figure}[t]
\centering
\includegraphics[width=\linewidth]{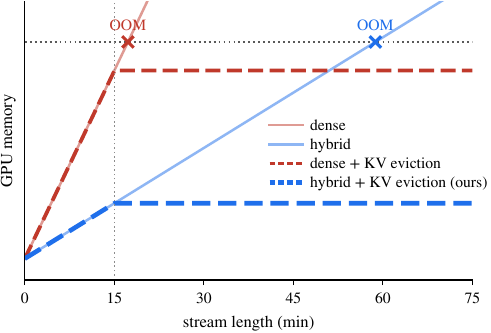}
\caption{
\textbf{KV cache growth with stream length.}
The hybrid backbone accumulates KV memory much more slowly than a dense model and needs far less memory for the same token budget.
Its full-attention cache still grows with stream length, however, so eviction is required for bounded memory.
}
\label{fig:teaser}
\end{figure}

Streaming video understanding requires processing an unbounded input stream under a bounded memory budget. Recent open-source language and multimodal backbones increasingly interleave linear and full attention layers~\cite{qwen35,kimilinear,minimax01}, introducing two distinct forms of temporal memory: a fixed-size recurrent state in linear attention and a token-wise KV cache in full attention. This distinction confines memory growth in streaming inference to the full-attention layers. In the 3:1 hybrid backbone used in this work~\cite{qwen35}, the recurrent states of the linear-attention layers remain fixed at 48 MiB, whereas the KV cache of the full-attention layers grows linearly with stream length, exceeding 10 GiB after one hour. Thus, despite substantially reducing the number of layers with growing memory, hybrid architectures still require explicit KV cache management to support bounded-memory streaming, as illustrated in Fig.~\ref{fig:teaser}. Moreover, eviction must occur before the question arrives, so the retention signal must be query-agnostic~\cite{infinipotv,streammem}.

Designed for dense models, where the KV cache is the only memory, existing KV eviction methods estimate which cached tokens are worth retaining from information within the KV cache itself, such as position, attention, or key-value representations~\cite{pitfalls}. Attention-based scores further require extra attention computation, often with proxy queries standing in for the unavailable question. All of these methods remain applicable to hybrid backbones through their full-attention layers. Beyond the KV cache, however, hybrid backbones offer a built-in source of information that continuously summarizes the preceding context within a fixed size and over multiple temporal scales: the recurrent state of linear-attention layers.

This state is already computed during the forward pass, so it can serve as a candidate retention signal without proxy queries or additional attention computation. In particular, gated-delta linear attention is designed to write into the state only the residual between the incoming representation and what can already be retrieved from the current state~\cite{deltanet,gateddeltanet}. Hence, how much a chunk changes the state reflects how much information it brings beyond what the state already holds~\cite{hola,dla}. We carry this criterion over to KV retention and introduce DeltaS, which scores each video chunk by its normalized state change, or \emph{state drift}, and prioritizes chunks with higher drift, \ie, those bringing more new information, for retention in the full-attention KV cache.

We evaluate DeltaS on Qwen3.5~\cite{qwen35}, an open-source multimodal hybrid model, across six long-video benchmarks under two fixed KV cache budgets. DeltaS achieves the best performance among query-agnostic, training-free, bounded-memory methods on all six benchmarks with an 8,192-token budget and on five of six with a 16,384-token budget, outperforming the strongest baseline by 2.1 and 1.7 points on average, respectively. Its advantage generally grows as the retained fraction of the video decreases. Our contributions are summarized as follows.

\begin{itemize}[leftmargin=1.2em,itemsep=2pt,topsep=3pt,parsep=0pt]

\item \textbf{State drift as a KV retention signal.} We propose DeltaS, which decides KV retention by the normalized state change. It requires no query or training, and its signal costs 1.9\% of the forward pass.

\item \textbf{Gains under tight memory budgets.} DeltaS outperforms existing query-agnostic, bounded-memory methods across six long-video benchmarks, with larger gains under tighter memory.

\item \textbf{Controlled comparison of retention signals.} With the budget and retention policy fixed, state drift outperforms position-, attention-, and key-value-based signals.

\end{itemize}

%% file: sec/2_related.tex
\section{Related Work}
\label{sec:related}

\paragraph{Bounded-memory streaming video.}
Approaches to memory-efficient streaming video understanding can be broadly grouped into training-based and inference-time methods. Training-based methods optimize models specifically for streaming~\cite{streamingvlm,streamforest,vst} or distill long-context capabilities into more efficient backbones~\cite{infinitevl}. Inference-time methods manage memory in a fixed pretrained model by compressing stored representations or selecting tokens to retain. Representation-based approaches reduce KV precision~\cite{kivi,streamingtom} or summarize earlier context into compact representations~\cite{protokv}. Token selection can occur before the question arrives or during query-time retrieval. ReKV, LiveVLM, and StreamKV retrieve relevant KV entries after observing the question~\cite{rekv,livevlm,streamkv}; the latter two also perform query-agnostic compression during streaming. Methods such as InfiniPot-V, StreamMem, and HERMES maintain a fixed memory budget by deciding what to retain before the question is known~\cite{infinipotv,streammem,hermes}. DeltaS targets this query-agnostic, training-free, bounded-memory setting.

\paragraph{Signals for KV cache eviction.} KV eviction methods derive retention scores from token position, attention, key or value representations, or combinations thereof~\cite{pitfalls}. Position-based strategies retain initial attention sinks and recent tokens~\cite{streamingllm,streamingvlm}, or sample tokens uniformly over time. Attention-based methods estimate importance from the attention received by cached tokens~\cite{h2o}, often using an observation window of recent tokens to compute these scores~\cite{snapkv}. In streaming video, StreamMem and HERMES use attention from chat-template tokens and generic guidance prompts, respectively, to guide retention~\cite{streammem,hermes}. Key-value-based methods instead use properties such as key norms~\cite{l2compress} or geometric distinctiveness based on cosine similarity to the mean key direction~\cite{keydiff}. InfiniPot-V compares keys at corresponding spatial positions with those in recent frames to identify temporal redundancy, and uses value norms to estimate the semantic importance of individual visual tokens~\cite{infinipotv}. Despite these differences, these methods derive their retention signals from information associated with the KV cache itself.

\paragraph{Recurrent state as a retention signal.}
Hybrid backbones maintain a fixed-size recurrent state alongside the KV cache~\cite{qwen35,kimilinear,minimax01}. Other approaches construct auxiliary memory, such as the temporal KV cache in StateKV~\cite{statekv} and the prototype-based summary state in ProtoKV~\cite{protokv}. Prediction error also guides memory retention and updates. B'MOJO retains unpredictable inputs in an explicit memory~\cite{bmojo}, while Titans updates a neural memory using gradient-based surprise~\cite{titans}. More directly, HOLA augments Gated DeltaNet with an exact KV cache, selects entries by delta-rule write magnitude, and trains the resulting architecture~\cite{hola}. DLA uses normalized state differences to determine memory-state boundaries and merges states to maintain a fixed capacity~\cite{dla}. Query-conditioned token reduction has also been studied in hybrid vision-language models to accelerate prefill~\cite{statefultokenreduction}. DeltaS instead reads changes in the existing recurrent state of a gated-delta hybrid backbone to guide KV retention in full-attention layers, without introducing a new memory module, additional training, or access to the downstream query.

%% file: sec/3_method.tex
\section{Method}
\label{sec:method}

\subsection{Streaming Setup and Hybrid Memory}
\label{sec:prelim}

We process streaming video in chunks as they arrive. Questions may arrive at any point in the stream, so preceding cache retention decisions must be made without access to the query. To bound memory usage, we limit the KV cache of each full-attention layer to at most $M$ retained tokens.

We use Qwen3.5-9B~\cite{qwen35} as our backbone, which interleaves one full-attention layer with every three Gated DeltaNet~\cite{gateddeltanet} linear-attention layers. Linear attention accumulates past context in a fixed-size recurrent state, whereas full attention stores a key-value pair for each token. The recurrent states across all linear-attention layers occupy a constant 48 MiB, while the KV caches across all full-attention layers grow by a total of 32 KiB per token. Bounding this growing memory therefore requires eviction from the full-attention KV caches.

\suppressfloats[t]
\begin{figure}[t]
\centering
\includegraphics[width=\linewidth]{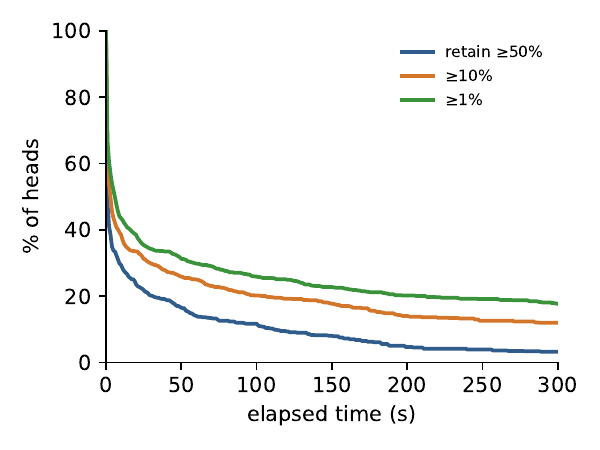}
\caption{\textbf{Distribution of state-retention horizons across linear-attention heads.} Each curve shows the percentage of heads retaining at least 50\%, 10\%, or 1\% of an initial state contribution over time. After 60 seconds, 14.1\% of heads still retain at least half of that contribution. The variation in retention horizons indicates that the recurrent state operates over multiple temporal scales.}
\label{fig:timescales}
\end{figure}

Importantly, the recurrent state does not operate at a single temporal scale. As shown in Fig.~\ref{fig:timescales}, some heads retain earlier state contributions for only a few seconds, whereas others preserve more than half of an earlier contribution after one minute. Heads with short and long retention horizons coexist across layers, and the ranking of per-head retention horizons is nearly identical across three benchmarks ($\rho \approx 0.998$). These observations suggest that the recurrent state captures past context over multiple temporal scales, providing a reference against which incoming chunks can be evaluated.

\subsection{State Drift}
\label{sec:drift}

\begin{figure*}[t]
\centering
\includegraphics[width=\textwidth]{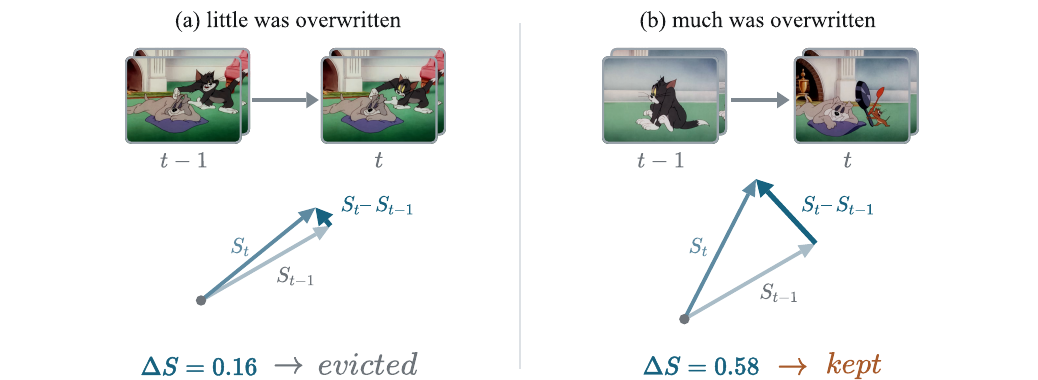}
\caption{\textbf{State drift as a KV retention signal.} In (a), visually similar consecutive chunks induce a small change from $S_{t-1}$ to $S_t$ and a low $\Delta S$. In (b), a larger visual change is accompanied by a larger state change and a higher $\Delta S$. The corresponding KV entries are evicted in (a) and retained in (b), illustrating the preference for chunks with larger state drift.}
\label{fig:delta}
\end{figure*}

We next examine how the recurrent state changes when a new input is processed. For a single linear-attention head, the gated delta rule~\cite{gateddeltanet} updates the state as follows, with layer and head indices omitted:
\begin{equation}
\label{eq:gdn}
S_t = S_{t-1}\bigl(\alpha_t (I - \beta_t k_t k_t^{\top})\bigr)
+ \beta_t v_t k_t^{\top},
\end{equation}
where $t$ indexes tokens, $k_t$ is the L2-normalized key vector, $v_t$ is the value vector, $\alpha_t$ controls state decay, and $\beta_t$ determines the write strength. Rearranging Eq.~\ref{eq:gdn} gives
\begin{equation}
\label{eq:gdn-expanded}
S_t = \alpha_t S_{t-1}
+ \beta_t \bigl(v_t - \alpha_t S_{t-1} k_t\bigr) k_t^{\top}.
\end{equation}
The update therefore depends on the residual between the incoming value $v_t$ and the value retrieved from the decayed state, $\alpha_t S_{t-1}k_t$. This residual is scaled by the write gate $\beta_t$ before being written to the state, so the write magnitude depends on both the prediction error and the write strength. The net change after processing a chunk reflects the combined effect of these writes and state decay over the chunk. Similar properties have been exploited in learned memory architectures, where delta-rule residuals or unexplained inputs are used to determine what should be memorized~\cite{bmojo,hola}.

Motivated by this update structure, we compare the recurrent states before and after processing each video chunk and use the resulting net change to score the chunk for retention. Denoting these states by $S_{\text{before}}$ and $S_{\text{after}}$, we define
\begin{equation}
\label{eq:deltas}
\Delta S =
\frac{1}{L}\sum_{i=1}^{L}
\frac{
    \left\lVert S^{i}_{\text{after}} - S^{i}_{\text{before}} \right\rVert_F
}{
    \left\lVert S^{i}_{\text{before}} \right\rVert_F
},
\end{equation}
where $L$ is the number of linear-attention layers and $\lVert\cdot\rVert_F$ denotes the Frobenius norm over the full state of each layer, including all heads. We refer to this normalized state change as \emph{state drift}. Normalizing by the pre-chunk state magnitude makes the change comparable across layers, and averaging yields a single score for each video chunk. Because the recurrent state summarizes preceding context, state drift measures how much an incoming chunk changes the state relative to that accumulated context, as illustrated in Fig.~\ref{fig:delta}. The state combines heads with different retention horizons, so this reference spans multiple temporal scales. Similar normalized state differences have also been used to define memory boundaries in linear-attention models~\cite{dla}.

Computing state drift adds little overhead to the standard forward pass. Both $S_{\text{before}}$ and $S_{\text{after}}$ are already available during inference, so Eq.~\ref{eq:deltas} requires only the norm of their difference and the norm of the previous state, without additional attention computation or an extra forward pass. We temporarily retain the previous state for this computation. Since the recurrent state has a fixed size, neither the storage required for this copy nor the cost of computing state drift grows with stream length. We report the measured runtime overhead in Sec.~\ref{sec:efficiency}.

\subsection{DeltaS: State-Guided KV Eviction}
\label{sec:policy}

\begin{figure*}[!t]
\centering
\includegraphics[width=0.95\textwidth]{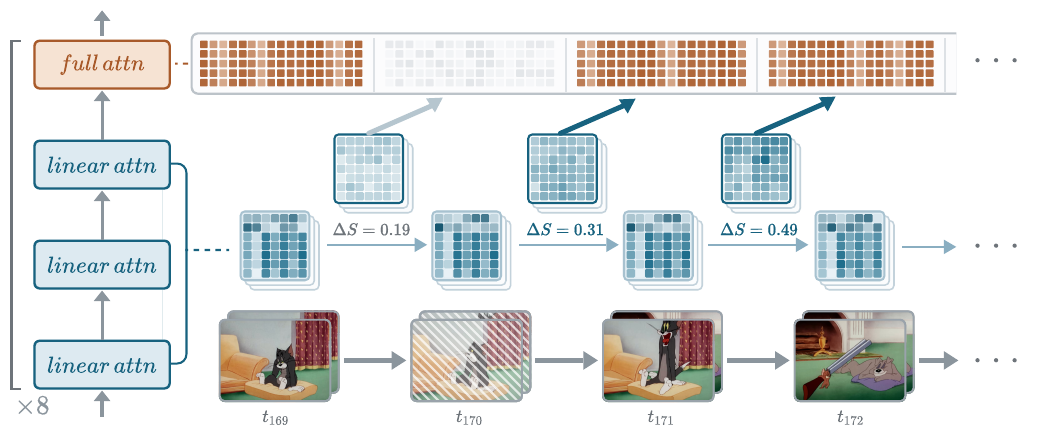}
\caption{\textbf{Overview of DeltaS.} Each video chunk updates the recurrent linear-attention states, whose normalized changes yield a shared score $\Delta S$ for retaining the corresponding KV entries across full-attention layers. KV blocks are vertically aligned with their source chunks. In the illustrated sequence, $t_{170}$ closely resembles the preceding chunk and has low state drift, whereas the more visually distinct chunk at $t_{172}$ has higher drift. Their corresponding KV entries are evicted (gray) and retained (orange), respectively.}
\label{fig:method}
\end{figure*}

DeltaS uses chunk-level state drift scores to guide KV retention. The score $\Delta S$ computed for each chunk is assigned to its tokens and guides retention of their corresponding KV entries across full-attention layers (Fig.~\ref{fig:method}). Later eviction steps reuse the scores assigned when each chunk was processed. Because this score aggregates recurrent-state changes across linear-attention layers, all full-attention layers use the same retention ranking. Each chunk corresponds to one second of video in our experimental setup (Sec.~\ref{sec:setup}).

We preserve two regions of the cache independently of the state drift score. The first $n_s$ tokens serve as attention sinks, while the most recent $n_w$ tokens provide a local context window~\cite{streamingllm,streamingvlm}. Once outside the recent window, non-sink tokens become candidates for score-based retention. When the cache exceeds the budget $M$, state drift determines which candidates are retained within the remaining capacity of $M-n_s-n_w$ tokens.

State drift scores tend to be higher near the beginning of the stream, so selecting tokens by a global ranking can concentrate retention in early portions of the video. To mitigate this bias, we divide candidate tokens into $N$ equal temporal spans based on their chunk arrival times and allocate an equal share of the remaining budget to each span. Within each span, tokens with higher $\Delta S$ are retained first. If a span contains fewer candidates than its quota, unused capacity is filled with the highest-scoring remaining candidates. The spans are recomputed at each eviction, requiring no advance knowledge of the final stream length. Using original arrival times rather than indices in the surviving cache keeps the partition based on elapsed time rather than token counts, even after repeated eviction.

Retained tokens keep their original position indices after eviction. We do not reassign consecutive positions to the surviving tokens, preserving both the original positional spacing between chunks and the positional information encoded in the cached keys. This eliminates the need to re-rotate keys after eviction.

To prevent question answering from altering the subsequent streaming state, we save the current state before processing each question. The model answers using the retained video KV and recurrent state, after which we restore the streaming state, including both memories, to its pre-question configuration. This also removes the KV entries generated during the question and response. Retention scores and chunk indices are restored along with the cache. Changes made during question answering therefore do not carry over to subsequent video processing or retention decisions, preserving the query-agnostic behavior of DeltaS.

Alg.~\ref{alg:deltas} summarizes the complete procedure.

\input{alg/alg_main}

%% file: alg/alg_main.tex
\begin{algorithm}[t]
\small
\caption{DeltaS: state-guided KV eviction}
\label{alg:deltas}
\begin{algorithmic}[1]
\Require budget $M$; sink $n_s$; window $n_w$; buckets $N$
\While{a new chunk $c$ arrives}
  \State $S_{\text{before}} \gets S$
  \State $\textsc{Prefill}(c)$ \Comment{updates $S$, appends $c$ to KV}
  \State $\Delta S \gets \operatorname{mean}_i \big( \lVert S^i - S^i_{\text{before}} \rVert_F / \lVert S^i_{\text{before}} \rVert_F \big)$
  \State $\mathrm{score}[t] \gets \Delta S$ \; for all tokens $t$ of $c$
  \If{$|\mathrm{KV}| > M$}
    \State $I_{\text{sink}} \gets$ first $n_s$, \; $I_{\text{window}} \gets$ last $n_w$
    \State $\mathcal{B}_1 \dots \mathcal{B}_N \gets$ $N$ equal time spans of the rest
    \State $I_{\Delta S} \gets$ top-$\big((M \!-\! n_s \!-\! n_w)/N\big)$ of each $\mathcal{B}_j$ by score
    \State $\mathrm{KV} \gets \mathrm{KV}[\,I_{\text{sink}} \cup I_{\Delta S} \cup I_{\text{window}}\,]$
  \EndIf
\EndWhile
\end{algorithmic}
\end{algorithm}

%% file: sec/4_experiments.tex
\section{Experiments}
\label{sec:experiments}

\subsection{Experimental Setup}
\label{sec:setup}

\input{tables/table1_body}

\paragraph{Backbone and streaming configuration.}
We evaluate all methods on the same frozen Qwen3.5-9B backbone~\cite{qwen35}. The model contains 24 linear-attention and eight full-attention layers, and KV eviction is applied only to the full-attention layers. For all methods, we sample videos at 2 fps and determine frame resolution using the fixed pixel-budget setting of StreamingVLM~\cite{streamingvlm}. DeltaS processes two frames per chunk, so each streaming step covers one second of video. Each chunk contains 98--112 visual tokens, depending on resolution, and 20 timestamp tokens.

We evaluate two per-layer KV budgets, $M=8{,}192$ and $M=16{,}384$, corresponding to approximately 256 MiB and 512 MiB of KV memory across the eight full-attention layers. At our sampling rate and tokenization settings, these budgets accommodate slightly over one and two minutes of video, respectively. For both budgets, DeltaS reserves $n_s=4$ sink tokens and a recent window of $n_w=1{,}024$ tokens, and uses $N=4$ temporal buckets. These settings are shared across all six benchmarks.

\paragraph{Benchmarks.}
We evaluate long-term context retention on six long-video understanding benchmarks: MLVU~\cite{mlvu}, Video-MME~\cite{videomme}, Video-MME v2~\cite{videommev2}, LongVideoBench~\cite{longvideobench}, LVBench~\cite{lvbench}, and EgoSchema~\cite{egoschema}. Each video is prefilled chronologically, chunk by chunk, and the question is provided only after the final chunk has been processed. Video-MME and LongVideoBench are evaluated without subtitles, and EgoSchema uses the public 500-item subset. We report accuracy (\%), with differences in percentage points, and detail splits and aggregation in Appendix~\ref{app:ports}. Averages, differences, and percentages are computed before rounding.

Results on StreamingBench~\cite{streamingbench} and OVO-Bench~\cite{ovobench} are reported in Appendix~\ref{app:streaming}. In our evaluation, a memoryless baseline using only recent frames outperforms Full KV on several tasks. These benchmarks do not consistently reward longer context, as also observed by SimpleStream~\cite{simplestream}. Aggregate scores alone therefore do not isolate long-term retention, so we report short-context baselines and separately analyze tasks requiring earlier context.

\paragraph{Baselines.}
We compare against three query-agnostic, training-free, bounded-memory methods---InfiniPot-V~\cite{infinipotv}, StreamMem~\cite{streammem}, and HERMES~\cite{hermes}---together with random, recency, and uniform retention controls. All bounded-memory methods receive the same KV budget $M$. We preserve each method's original design choices for chunk size, reserved regions, budget allocation, and head-wise selection, adapting only components that cannot be applied directly because of architectural differences in the hybrid backbone. Method-specific settings and adaptations are detailed in Appendix~\ref{app:ports}.

We additionally include Full KV and ReKV as reference settings. \emph{Full KV} follows the same streaming inference path with eviction disabled, so its KV memory grows with video length. ReKV~\cite{rekv} retrieves relevant KV entries after the question arrives and is therefore query-aware.

\subsection{Main Results}
\label{sec:main}

Table~\ref{tab:main-vmme2} shows that DeltaS consistently performs well under bounded-memory streaming. At $M=8{,}192$, it achieves the best results among query-agnostic, bounded-memory methods on all six benchmarks, outperforming StreamMem, the strongest baseline on average, by 2.1 points. At $M=16{,}384$, it remains best on five of six benchmarks, with an average gain of 1.7 points. Notably, DeltaS at $M=8{,}192$ achieves an average accuracy of 58.7, nearly matching StreamMem at $M=16{,}384$ (58.8), despite using half the KV budget.

The advantage is particularly pronounced when the fixed cache budget can retain only a small fraction of the video. On LVBench, the longest benchmark in our evaluation, DeltaS exceeds the strongest bounded-memory baseline by 5.6 points at $M=8{,}192$ and 4.3 points at $M=16{,}384$. Its 46.0 at $M=8{,}192$ also surpasses StreamMem's 44.1 at twice the budget. In contrast, differences among methods are smaller on shorter videos such as EgoSchema, where a substantial fraction of the video fits within the cache.

\begin{figure}[t]
\centering
\includegraphics[width=\linewidth]{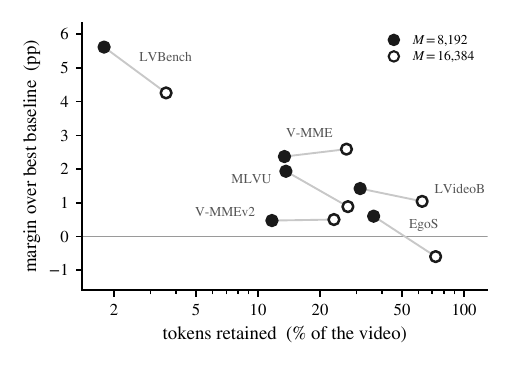}
\caption{\textbf{Performance margin as a function of KV retention rate.} Each point shows DeltaS's accuracy margin over the best query-agnostic, bounded-memory baseline for a benchmark and budget. Lines connect the two budgets for each benchmark. DeltaS generally shows larger gains at lower retention rates.}
\label{fig:retention}
\end{figure}

Figure~\ref{fig:retention} plots DeltaS's margin over the strongest bounded-memory baseline against retention rates calculated using each benchmark's median video length and the corresponding KV budget. Because video lengths vary across benchmarks, the same budget of $M=8{,}192$ yields retention rates ranging from 1.8\% on LVBench to 36.4\% on EgoSchema. DeltaS generally shows larger gains at lower retention rates, suggesting that retention choices become more consequential when only a small fraction of a long stream can remain in the KV cache. Video-MME v2 exhibits a different pattern: DeltaS maintains an approximately 0.5-point advantage over StreamMem at both budgets, with no widening of the gap under the tighter budget. Both methods gain approximately 1.5--1.6 points when the budget doubles. The benchmark's demanding evidence integration and temporal reasoning tasks~\cite{videommev2} may limit the extent to which differences in retention translate into accuracy gains as memory pressure increases. Full KV also reaches only 32.8 under our evaluation setup, consistent with challenges beyond KV retention.

These comparisons evaluate complete eviction systems, which differ in both scoring signals and retention policies. To examine whether state drift provides an informative signal for KV retention, we next compare it with alternative signals under a shared policy and memory budget.

\subsection{Controlled Comparison of Retention Signals}
\label{sec:signals}

To assess whether state drift provides informative cues for KV retention, Table~\ref{tab:score-2col} compares representative signals under a common retention setup. The comparison covers the categories introduced in Sec.~\ref{sec:related}: temporal position (uniform), attention (SnapKV~\cite{snapkv}), and key- or value-based representations (L2-norm~\cite{l2compress}, KeyDiff~\cite{keydiff}, and the temporal-axis redundancy (TaR) and value-norm (VaN) scores of InfiniPot-V~\cite{infinipotv}).

We additionally use HOLA's scoring function, $\beta\lVert e\rVert$~\cite{hola}, as a state-based baseline, where $e$ is the prediction residual and $\beta$ controls the write strength. We apply this scoring rule to the frozen backbone without adopting HOLA's memory architecture. This signal measures the magnitude of the gated-delta write, whereas DeltaS measures the normalized net change in the state before and after processing a chunk. Both signals assign a single score to each chunk, so the two are compared at the same scoring granularity.

All signals are evaluated with $M=8{,}192$, four sink tokens, a recent window of 1,024 tokens, and two-frame input chunks. Score-based methods use the same temporal-bucket retention policy, while uniform samples evenly over time. We preserve conventions intrinsic to each signal, including scoring granularity and KV-head aggregation; the configurations are detailed in Appendix~\ref{app:signals}.

\input{tables/table2_body}

Table~\ref{tab:score-2col} shows that both state-based signals achieve higher accuracy than the position-, attention-, and key-value-based alternatives on LongVideoBench, MLVU, and LVBench. DeltaS ranks first on all three benchmarks, exceeding VaN, the strongest cache-internal signal, by 0.7--2.1 points. It also outperforms the state-based baseline $\beta\lVert e\rVert$ by 0.5--1.2 points.

These results support the view that the recurrent state carries useful information for KV retention. DeltaS accesses this information through state changes already produced by the backbone, demonstrating their utility for deciding which content to retain under a fixed memory budget.

\begin{figure}[t]
\centering
\includegraphics[width=0.9\linewidth]{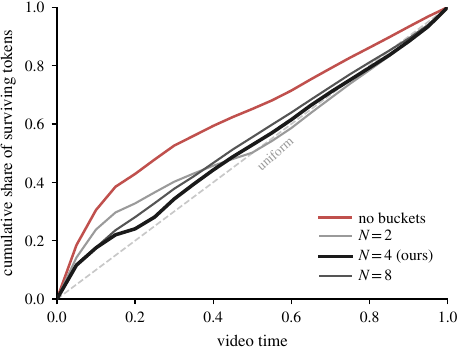}
\caption{\textbf{Effect of temporal bucketing on retained-token distribution.} Curves show the cumulative share of retained tokens over normalized video time; the dashed diagonal indicates uniform temporal coverage. Temporal bucketing reduces the early concentration observed without buckets.}
\label{fig:bucket}
\end{figure}

\subsection{Effect of Temporal Bucketing}

As discussed in Sec.~\ref{sec:policy}, global ranking by $\Delta S$ tends to favor earlier parts of the stream. Figure~\ref{fig:bucket} shows the cumulative share of retained tokens over normalized video time. Curves above the diagonal, which represents uniform temporal coverage, indicate a concentration of retained tokens toward the beginning of the video. This early bias is pronounced without bucketing. Temporal buckets bring the curves closer to the diagonal, distributing retained tokens more evenly across the video.

To assess whether this temporal redistribution also improves accuracy, we compare different bucket counts $N$ on LongVideoBench, MLVU, and LVBench. The total KV budget is held fixed across bucket counts. Since the recent window itself guarantees retention of the latest tokens, we exclude this reserved region in the ablation to examine the effect of temporal bucketing.

\begin{figure}[t]
\centering
\includegraphics[width=\linewidth]{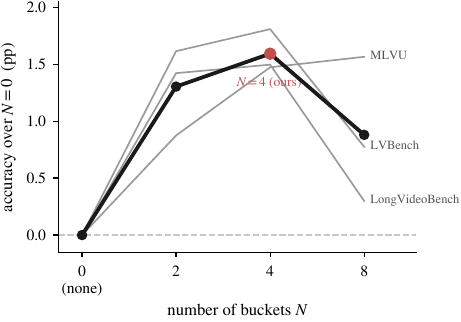}
\caption{\textbf{Effect of bucket count on accuracy.} Gains over the baseline without bucketing ($N=0$) are shown for LongVideoBench, MLVU, and LVBench. The thick black line shows the mean gain across the three benchmarks, which peaks at $N=4$ among the tested settings.}
\label{fig:bucket_performance}
\end{figure}

Figure~\ref{fig:bucket_performance} reports accuracy gains over the baseline without temporal bucketing, denoted by $N=0$ in the figure. The tested bucket configurations outperform this baseline on all three benchmarks. The average gain is largest at $N=4$ and decreases when $N$ is increased to eight. These results suggest that temporal bucketing can improve accuracy as well as temporal coverage, while preserving selection by state drift within each span.

\subsection{Efficiency Analysis}
\label{sec:efficiency}

In streaming inference, repeated KV eviction makes retention-signal computation a recurring cost. DeltaS uses recurrent states already produced by the forward pass, requiring a copy of the previous state and the norms of the state difference and the previous state. No additional forward pass or explicit attention computation is needed. In contrast, attention-based methods such as SnapKV, StreamMem, and HERMES perform additional attention computation or proxy-query processing to obtain their retention scores.

\begin{figure}[t]
\centering
\includegraphics[width=\linewidth]{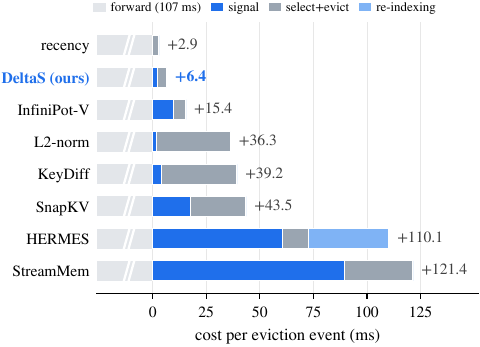}
\caption{\textbf{Runtime overhead per KV eviction event.} Colored segments show additional latency from signal computation, selection and eviction, and cache re-indexing. The light gray segment represents the reference 107 ms chunk forward time, compressed by the axis break. The total additional cost of $\beta\lVert e\rVert$ (206.1 ms) is omitted for readability.}
\label{fig:cost}
\end{figure}

We measure runtime on a single A100 by streaming a 4,730-second MLVU video with $M=8{,}192$ for 600 steps. For runtime measurements, all methods use two-frame chunks, and we report the median latency over steps in which eviction occurs for each method. Computing the state drift signal takes 2.07 ms, corresponding to 1.9\% of the reference chunk forward time of 107 ms. Including selection and eviction, the total additional cost is 6.4 ms per eviction event, the lowest among the evaluated methods that compute a retention signal (Fig.~\ref{fig:cost}). Under the same measurement protocol, InfiniPot-V adds 15.4 ms. The additional latency of StreamMem and HERMES exceeds the chunk forward time itself.

The comparison with $\beta\lVert e\rVert$ shows that state-based signals can differ substantially in computational cost. In our implementation, computing token-level residuals for this signal brings its total additional cost to 206.1 ms per eviction event, including selection and eviction. DeltaS avoids this computation by scoring chunks from the states already available before and after processing, achieving higher accuracy in the controlled comparison at lower cost.

%% file: tables/table1_body.tex
\begin{table*}[t]
\renewcommand{\band}[1]{\multicolumn{8}{@{\hspace{4pt}}l}{\textbf{#1}}\\[1pt]}%
\renewcommand{\hd}[1]{\makebox[48pt]{\makecell{#1}}}%
\renewcommand{\hda}[1]{\makebox[48pt]{#1}}%
\renewcommand{\hdm}[1]{\makebox[73pt][l]{#1}}%
\centering
\footnotesize
\setlength{\tabcolsep}{4pt}
\sbox{\mtbox}{%
\begin{tabular}{@{\hspace{4pt}}l @{\hspace{4.4pt}} cccccc c@{\hspace{4pt}}}
\toprule
\hdm{Method} & \hd{MLVU} & \hd{Video-\\MME} & \hd{Video-\\MME v2}
       & \hd{Long\\VideoBench} & \hd{LVBench}
       & \hd{Ego\\Schema} & \hda{Avg} \\
\midrule
\band{Unbounded or query-aware}
\m{Full KV} & 74.9 & 74.5 &  32.8  & 70.2 & 55.1 & 71.8 &  63.2 \\
\m{ReKV} & 71.1 / 73.2 & 69.7 / 71.1 &  29.7 / 31.7  & 66.0 / 68.1 & 50.6 / 52.4 & 71.4 / 72.4 &  59.8 / 61.5 \\
\midrule
\band{Simple baselines}
\m{random} & 59.7 / 63.0 & 63.1 / 67.7 &  25.2 / 27.9  & 59.8 / 62.5 & 38.4 / 40.3 & \underline{71.2} / \underline{71.4} &  52.9 / 55.5 \\
\m{recency} & 58.7 / 62.0 & 61.8 / 65.8 &  23.8 / 25.6  & 56.1 / 60.9 & 36.9 / 36.9 & 68.8 / 70.8 &  51.0 / 53.7 \\
\m{uniform} & 59.4 / 62.8 & 63.0 / 66.8 &  26.2 / 27.3  & 58.7 / 61.9 & 37.9 / 40.1 & 69.0 / \textbf{72.0} &  52.4 / 55.2 \\
\band{Streaming KV compression}
\m{InfiniPot-V} & 62.9 / 67.8 & 65.4 / 68.7 &  26.7 / 29.2  & 60.3 / 64.2 & 38.9 / 41.0 & 70.0 / \underline{71.4} &  54.0 / 57.0 \\
\m{StreamMem} & \underline{68.7} / \underline{72.0} & \underline{67.4} / 69.1 &  \underline{29.6} / \underline{31.2}  & \underline{62.8} / \underline{65.4} & \underline{40.4} / \underline{44.1} & 71.0 / 71.0 &  \underline{56.6} / \underline{58.8} \\
\m{HERMES} & 65.1 / 68.9 & 66.1 / \underline{69.4} &  26.9 / 28.5  & 61.5 / 63.4 & 37.1 / 37.8 & 70.8 / \textbf{72.0} &  54.6 / 56.7 \\
\rowcolor{oursbg}[4pt][4pt]
\m{\textbf{DeltaS (ours)}} & \textbf{70.6} / \textbf{72.9} & \textbf{69.7} / \textbf{72.0} &  \textbf{30.1} / \textbf{31.7}  & \textbf{64.2} / \textbf{66.5} & \textbf{46.0} / \textbf{48.4} & \textbf{71.8} / \underline{71.4} &  \textbf{58.7} / \textbf{60.5} \\
\bottomrule
\end{tabular}}%
\usebox{\mtbox}
\caption{\textbf{Performance on six long-video benchmarks under two KV cache budgets.} Accuracy (\%) is reported for M=8,192/16,384. All methods are evaluated on the same Qwen3.5-9B backbone, while method-specific conventions for reserved regions, chunk size, budget allocation, and KV heads follow their respective implementations (Appendix~\ref{app:ports}). Full KV and ReKV are included as unbounded-memory and query-aware reference methods, respectively, and are excluded from ranking. Best and second-best results among query-agnostic, bounded-memory methods are shown in \textbf{bold} and \underline{underline} for each budget.}
\label{tab:main-vmme2}
\end{table*}

%% file: tables/table2_body.tex
\begin{table}[t]
\centering
\footnotesize
\renewcommand{\band}[1]{\multicolumn{4}{l}{\textbf{#1}}\\[1pt]}%
\renewcommand{\hd}[1]{\makebox[32pt]{\makecell{#1}}}%
\renewcommand{\hscore}[1]{\makebox[92pt][l]{#1}}%
\providecommand{\hdwB}[1]{\makebox[45pt]{\makecell{#1}}}%
\setlength{\tabcolsep}{4pt}
\begin{tabular}{l @{\hspace{8.4pt}} ccc}
\toprule
\hscore{Method} & \hdwB{Long\\VideoBench} & \hd{MLVU} & \hd{LVBench} \\
\midrule
\band{Position}
\m{uniform} & 58.7 & 59.4 & 37.9 \\
\midrule
\band{Cache-internal signals}
\m{SnapKV~\cite{snapkv}} & 62.0 & 66.4 & 40.3 \\
\m{L2-norm~\cite{l2compress}} & 60.8 & 63.6 & 36.2 \\
\m{KeyDiff~\cite{keydiff}} & 61.2 & 67.1 & 39.8 \\
\m{TaR~\cite{infinipotv}} & 62.4 & 68.7 & 40.1 \\
\m{VaN~\cite{infinipotv}} & 63.4 & 68.9 & 44.0 \\
\midrule
\band{State-based signals}
\m{$\beta\lVert e\rVert$~\cite{hola}} & \underline{63.6} & \underline{69.4} & \underline{44.9} \\
\ocL{\m{DeltaS \,\scriptsize(ours)}} & \oc \textbf{64.2} & \oc \textbf{70.6} & \oc \textbf{46.0} \\
\bottomrule
\end{tabular}
\caption{\textbf{Controlled comparison of KV retention signals.} All methods use the same memory budget ($M=8{,}192$) and reserved regions. Score-based methods share a temporal-bucket retention policy, while uniform samples evenly over time. Scoring granularity and KV-head aggregation follow each signal's definition (Appendix~\ref{app:signals}).}
\label{tab:score-2col}
\end{table}

%% file: sec/5_conclusion.tex
\section{Conclusion}
\label{sec:conclusion}

We introduced DeltaS, a query-agnostic and training-free KV eviction method that uses recurrent-state drift in gated-delta hybrid video-language models as a retention signal.
Across six long-video benchmarks, DeltaS achieves strong performance among bounded-memory baselines, with generally larger gains under tighter memory constraints.
Controlled comparisons under a shared memory budget and retention policy show that state-based signals yield higher accuracy than position-, attention-, and key-value-based alternatives, supporting the utility of recurrent-state changes for KV retention.
Temporal bucketing complements this signal by reducing the tendency to favor early parts of the stream and maintaining broader temporal coverage.
By reusing states already produced by the forward pass, DeltaS achieves these gains with little additional computation.
These results show that the fixed-size recurrent state can both summarize past context and guide which content to retain in the KV memory of full-attention layers.

\section{Limitations}
\label{sec:limitations}

Our evaluation uses a single hybrid backbone; the effectiveness of DeltaS across other hybrid architectures and model scales remains to be established. Our interpretation of state drift rests on the gated delta rule's residual update, and its utility as a retention signal under other state-update rules requires further validation. Finally, all tokens in a chunk share a single score, so DeltaS cannot distinguish more informative regions within a chunk through its scoring mechanism. Scoring at finer granularity is a natural extension.

%% file: sec/A_ports.tex
\section{Implementation Details}
\label{app:ports}

\subsection{Method Adaptations for Table~\ref{tab:main-vmme2}}

Table~\ref{tab:main-vmme2} compares complete streaming systems on a common Qwen3.5-9B backbone. Every method receives the same KV budget $M$, while method-specific choices such as chunk size, reserved regions, budget unit, and head-wise selection follow the corresponding original formulation whenever possible. We introduce substitutions only when architectural differences prevent a direct port; these cases are described below. Since only the eight full-attention layers maintain KV caches, eviction is applied exclusively to those layers. Table~\ref{app:tab:ports} summarizes the resulting configurations.

\begin{table*}[t]
\centering
\scriptsize
\setlength{\tabcolsep}{3.2pt}
\begin{tabular}{l l l l l l l l}
\toprule
\textbf{Method} & \textbf{Ref. impl.} & \textbf{Chunk} & \textbf{Always kept} & \textbf{Recent region} & \textbf{Budget unit} & \textbf{Head axis} & \textbf{Period} \\
\midrule
random & --- & 2 frames & sink 4 & 1{,}024 & token & shared & every step \\
recency & --- & 2 frames & sink 4 & $M-4$ & token & shared & every step \\
uniform & --- & 2 frames & sink 4 & 1{,}024 & token & shared & every step \\
InfiniPot-V & author code & 2 frames & text prefix & --- (query frames) & frame & per KV head & every 25\% of $M$ \\
StreamMem & paper & 8 frames & text prefix & --- (from the score) & token & shared & every step \\
HERMES & author code & 16 frames & text prefix & --- (from the score) & token & shared & every step \\
\textbf{DeltaS (ours)} & --- & 2 frames & sink 4 & 1{,}024 & token & shared & every step \\
\emph{Full KV} & --- & 2 frames & --- & --- & --- & --- & no eviction \\
\emph{ReKV} & author code & 2 frames & preamble 16 & 512 (ingest) & token & shared & once at question time \\
\bottomrule
\end{tabular}
\caption{\textbf{Implementation conventions for Table~\ref{tab:main-vmme2}.} All bounded methods share the per-layer budget $M$; other conventions follow the original implementations. ``---'' in Recent region: no reserved region, with recent tokens retained through the method's score (for InfiniPot-V, query frames within the TaR budget). HERMES keeps $M-1$ tokens in deep layers plus one summary token.}
\label{app:tab:ports}
\end{table*}

\paragraph{Chunk size and compression schedule.}
Our default streaming path processes two frames per chunk. StreamMem and HERMES instead use the 8-frame and 16-frame chunks specified by their original formulations, respectively; forcing them to two frames would alter the methods rather than reproduce them. Compression is triggered whenever the cache exceeds its budget. This occurs at every streaming step for most methods. InfiniPot-V is an exception because each compression removes 25\% of its budget, requiring approximately 18.5 subsequent steps at $M=8{,}192$, and about twice as many at $M=16{,}384$, before the cache fills again.

\paragraph{InfiniPot-V.}
We follow the standard hyperparameters reported in the paper: half of the budget allocated to TaR and half to VaN, a recent-frame ratio of 0.125, and a compression rate of 0.75. The public implementation defaults to a recent-frame ratio of 0.25; we use the paper's value of 0.125 instead. For TaR, we follow the public implementation in operating on keys after positional encoding. Although the paper describes keys before positional encoding, the released code uses post-encoding keys, which also matches the representation stored by our backbone. This distinction affects only TaR; VaN operates on values and is therefore unchanged. Separator text inserted between video chunks is included among compressible tokens, preventing these tokens from accumulating outside the fixed budget on long streams.

\paragraph{StreamMem.}
Because no public implementation is available, we use the paper as the reference. The paper expresses its score as a $q \times n$ matrix without specifying a head axis, which we interpret as a score shared across heads. Two adaptations are required by our backbone. First, the temporal-grid step of two frames is used as the unit for the input-frame filter. Filtering individual frames would re-pair the surviving frames and thereby alter their representation, whereas filtering at the grid-step level preserves the original encoded units. Second, when duplicate frames are identified, we retain one representative rather than averaging them. All other hyperparameters follow the ablation-optimal configuration reported by the authors, including merging and a filter threshold of 0.95.

\paragraph{HERMES.}
We use the approximate-attention path selected by the released implementation under its default attention backend. When flash attention or SDPA is used, this is the path executed by the original code rather than exact attention. One adaptation is required for the layer hierarchy: directly applying the original thresholding rule to only eight full-attention layers leaves the shallow group empty. We therefore enforce at least one shallow layer, after which seven of the eight layers follow the original assignment and one differs. Middle-layer coefficients follow the paper. Summary-token handling and position re-indexing are ported together; after re-deriving the indexing expression for Qwen3.5, we verify that it is bit-identical to a direct transcription of the original operation.

\paragraph{Simple controls.}
The random, recency, and uniform baselines differ only in their retention rule. Random eviction samples at chunk granularity to avoid introducing token-level fragmentation as an additional variable. Recency allocates the entire non-sink budget to the most recent context, corresponding to approximately the latest 65 seconds at $M=8{,}192$. Uniform retention samples over absolute stream time rather than the indices currently present in the cache; sampling over surviving indices would progressively bias retention toward recent content after repeated eviction.

\paragraph{Reference settings.}
Full KV follows the same streaming path with eviction disabled and therefore provides an unbounded-memory reference. ReKV is query-aware: during streaming it retains the first 16 tokens corresponding to the original chat preamble together with the most recent 512 tokens, and after the question arrives it fills the available context using retrieval.

\paragraph{Verification against reference implementations.}
Because all methods are evaluated on a different backbone from their original reports, reproducing published accuracy values directly is not meaningful. We instead verify the equivalence of key implementation components where possible. HERMES position re-indexing is bit-identical to a transcription of the original implementation, and InfiniPot-V TaR scores are byte-identical to the reference computation when the complete frame grid is retained.

\subsection{Controlled Retention-Signal Comparison}
\label{app:signals}

Table~\ref{tab:score-2col} is designed to isolate the retention signal from the surrounding streaming system. All rows therefore share a budget of $M=8{,}192$, four sink tokens, a recent window of 1{,}024 tokens, and two-frame chunks. Eviction is applied independently to each full-attention layer whenever its cache exceeds the budget, and all scores are computed in float32. Signals whose original definitions assign greater importance to smaller values are sign-inverted so that all rows follow the same top-score retention convention.

Method-specific choices that are inseparable from the score itself, most notably head-wise aggregation and scoring granularity, are preserved. The uniform baseline is the only exception to the temporal bucket allocation used by the remaining rows, since uniform sampling already enforces coverage along time. Thus, Table~\ref{tab:score-2col} should be interpreted as a controlled comparison of retention signals rather than a reproduction of each complete method.

\begin{table*}[t]
\centering
\footnotesize
\setlength{\tabcolsep}{3pt}
\begin{tabular}{l l p{0.30\linewidth} l l l}
\toprule
\textbf{Signal} & \textbf{Reads} & \textbf{Scoring function} & \textbf{Head axis} & \textbf{Granularity} & \textbf{Reference impl.} \\
\midrule
uniform & position & even indices along time (no score) & --- & --- & --- \\
SnapKV & attention & column sum of observation-window attention $\rightarrow$ max pooling (kernel 7) & per KV head & token & paper Listing 1 \\
L2-norm & key & $-\lVert k\rVert_2$ & per KV head & token & author code \\
KeyDiff & key & $-\cos(k,\,\text{mean key direction})$ & per KV head & token & kvpress \\
TaR & key & negative mean cosine with recent keys at the same patch coordinate & per KV head & token & author code \\
VaN & value & $\lVert v\rVert_2$ & per KV head & token & author code \\
$\beta\lVert e\rVert$ & state & chunk mean of $\beta\lVert v-\alpha S k\rVert_2$ & shared & chunk scalar & paper Eq.~6 \\
DeltaS (ours) & state & layer mean of $\lVert S^{i}_{\text{after}} - S^{i}_{\text{before}}\rVert_F / \lVert S^{i}_{\text{before}}\rVert_F$ & shared & chunk scalar & --- \\
\bottomrule
\end{tabular}
\caption{\textbf{Retention-signal configurations for Table~\ref{tab:score-2col}.} The memory budget and reserved regions are fixed across rows. Head-wise aggregation and scoring granularity follow the corresponding signal definition where they cannot be separated from the score itself.}
\label{app:tab:signals}
\end{table*}

The signals are ordered by the information they use: position, attention, key or value representations, and recurrent state.

\paragraph{SnapKV.}
SnapKV uses an observation window to estimate the attention received by cached tokens. In the original single-prefill setting, this window consists of the final 32 prompt tokens. Because streaming has no fixed prompt endpoint, we instead use the last 32 tokens of the currently arriving chunk as the observation window and recompute the score at every streaming step. As in the original method, scores are not accumulated over time. Attention columns are summed and max-pooled with kernel size 7 along the cache-index axis. Although repeated eviction introduces gaps along this axis, the temporal ordering of surviving tokens is preserved. The original observation tokens are always retained; in our setting they fall within the common 1{,}024-token recent window.

\paragraph{L2-norm.}
The original method leaves the first two layers uncompressed because key norm is poorly correlated with attention in those layers. In Qwen3.5, layers 0 and 1 are Gated DeltaNet layers and do not maintain KV caches, so no additional exclusion is required. Scores are computed directly from the current key cache at eviction time without an accumulation buffer. Because the L2 norm is rotation-invariant, the score is unchanged by positional rotation.

\paragraph{KeyDiff.}
We use NVIDIA kvpress as the reference implementation because no author code is available. Following the formulation in the paper, keys are normalized before their mean direction is computed. Normalizing after averaging would yield a different anchor direction, although the paper reports negligible accuracy differences between the variants. The anchor is recomputed over the complete surviving cache at each eviction. Because positional encoding rotates each key by a position-dependent angle, cosine similarities between keys change, so we use keys after positional encoding, consistent with the reference implementation.

\paragraph{TaR.}
The original TaR formulation restores the video frame grid and compares keys at the same spatial patch coordinate. Token-level eviction destroys this complete grid, so for the controlled comparison we use an equivalent linearized computation: for each patch coordinate, we compute the mean key direction over the recent region and compare retained keys against it. When the full grid is available, this formulation produces byte-identical scores to a transcription of the original method. To maintain the common reserved-region setting of Table~\ref{tab:score-2col}, TaR uses our fixed recent region rather than a fraction of all frames. Text tokens are scored using the same formulation.

\paragraph{VaN.}
We retain the original score $\lVert v\rVert_2$ but disable the method's layer-adaptive pooling. The original pooling size is determined from the coefficient of variation of value norms over the frame grid, which is no longer well defined after token-level cache thinning. Values themselves do not receive positional encoding.

\paragraph{$\beta\lVert e\rVert$.}
We transfer only the scoring function from the learned memory architecture, without its architecture, training procedure, or cache design. The residual write $\beta_t e_t k_t^{\top}$ has rank at most one. For unit-norm keys and nonnegative $\beta_t$, its Frobenius norm is $\beta_t\lVert e_t\rVert_2$. We compute this quantity per head, average across heads and across the linear-attention layers, and then average across the tokens of each chunk, producing one shared chunk-level score for all full-attention layers. Matching the chunk granularity of DeltaS allows the two state-based scores to be compared without introducing a difference in selection granularity. We use the product $\beta\lVert e\rVert$, rather than either factor individually, following the ablation reported by the original authors.

\paragraph{Shared constraints in the controlled comparison.}
TaR and VaN jointly form the original InfiniPot-V system, where the available budget is divided between their two scores and the selected sets are combined with a recent region. In Table~\ref{tab:score-2col}, each signal instead ranks the entire candidate cache independently under the same budget, so these rows represent the individual signals rather than a reproduction of InfiniPot-V.

Similarly, SnapKV originally expands KV entries to query heads and performs selection per query head. Because Qwen3.5 stores the cache per KV head, we aggregate scores over query heads that share the same KV head before selection.

Finally, all signals are evaluated with the same four sink tokens, 1{,}024-token recent region, and token-count budget. Several original formulations use pure top-$k$ selection or frame-based budgets; those conventions are intentionally replaced here to create a common retention setting. Consequently, 12.5\% of the $M=8{,}192$ budget is always reserved for recent context, while the scoring signal determines retention within the remaining capacity. We do not equalize scoring granularity, since averaging a token-level score into chunk-level values would itself modify the signal's original behavior.

\subsection{Reproducibility Details}

\begin{table*}[t]
\centering
\footnotesize
\setlength{\tabcolsep}{3pt}
\begin{tabular}{l l r l}
\toprule
\textbf{Benchmark} & \textbf{Split} & \textbf{Items} & \textbf{Aggregation} \\
\midrule
MLVU & dev (MC) & 2{,}173 & M-Avg (mean over 7 tasks) \\
Video-MME & full & 2{,}700 & mean over items \\
Video-MME v2 & test & 3{,}200 & mean over items \\
LongVideoBench & val & 1{,}337 & mean over items \\
LVBench & full & 1{,}549 & mean over items \\
EgoSchema & subset & 500 & mean over items \\
StreamingBench & real-time (10 tasks) & 2{,}495 & mean over items (Appendix B) \\
OVO-Bench & MC regimes (9 of 12 tracks) & 1{,}468 & mean over items within regime (Appendix B) \\
\bottomrule
\end{tabular}
\caption{\textbf{Benchmark splits and evaluation protocols.} MLVU uses the official M-Avg metric. StreamingBench and OVO-Bench are analyzed separately in Appendix~\ref{app:streaming}.}
\label{app:tab:bench}
\end{table*}

\paragraph{Model and inference.}
We use the public Qwen3.5-9B checkpoint in bfloat16 without modifying its weights. The model contains 24 Gated DeltaNet layers and eight full-attention layers at indices 3, 7, 11, 15, 19, 23, 27, and 31. Each full-attention layer has four KV heads with head dimension 256, resulting in 32 KiB of KV memory per token across all eight layers. In contrast, the recurrent linear-attention state remains fixed in size independently of the number of processed tokens. We use the PyTorch linear-attention implementation provided by Transformers. The 107 ms chunk-forward time reported in Sec.~\ref{sec:efficiency} is measured under this configuration and is shared across all methods.

\paragraph{Video preprocessing.}
Videos are sampled uniformly at 2 fps. Frame resolution follows StreamingVLM's pixel-budget rule, using the total number of sampled frames. Each two-frame chunk contains 98--112 visual tokens depending on resolution, together with 20 timestamp tokens.

\paragraph{Prompt and streaming state.}
Video chunks are appended sequentially using the same chat template and timestamp convention for all methods. Timestamp text is relative to the corresponding chunk. When a question is answered, KV entries produced during the question and response are discarded afterward, and ingestion resumes from the cache state that preceded the question, as described in Sec.~\ref{sec:policy}. Thus, question turns do not modify the subsequent streaming state.

\paragraph{Benchmark protocols.}
Table~\ref{app:tab:bench} summarizes the benchmark splits, item counts, and aggregation rules. Video-MME and LongVideoBench are evaluated using frames only. Although the official LongVideoBench protocol includes interleaved subtitles, we omit subtitles so that both benchmarks are evaluated under the same visual-only condition. EgoSchema is evaluated on the public 500-item subset. On Video-MME v2, we report per-question accuracy rather than the official group-based non-linear score, so these values are not directly comparable to the official leaderboard. For the streaming benchmarks, we evaluate the real-time visual understanding track of StreamingBench and the two multiple-choice regimes of OVO-Bench, backward tracing and real-time visual perception; the forward active responding regime is excluded because it follows a dense-query protocol along the timeline rather than a single question after the observed prefix. OVO-Bench results are item-weighted within each regime, rather than averaging the per-track means as in the official scorer; the 68.6/67.7 pair reported in Appendix B is the only result computed using the official aggregation.

\paragraph{Memory budget.}
The budget $M$ denotes the number of tokens retained by each full-attention layer. We use $M=8{,}192$ and $M=16{,}384$, corresponding to approximately 256 MiB and 512 MiB of KV memory, or slightly over one and two minutes of video at our sampling rate. All eight full-attention layers receive the same budget.

\paragraph{Runtime measurement.}
The efficiency analysis in Sec.~\ref{sec:efficiency} streams a 4{,}730-second MLVU video for 600 steps on a single A100. Each method is measured in three repetitions; we discard the first as GPU warm-up and report the median, over the remaining repetitions, of the median latency across steps in which eviction occurs. For this measurement only, all methods use two-frame chunks so that the cost of a single eviction event is compared under the same input size; accuracy is not evaluated under this modified setting. The decomposition shown in Fig.~\ref{fig:cost} is obtained from instrumentation and should be interpreted as less precise than the measured total latency.

\paragraph{Hardware.}
Accuracy evaluations are run on NVIDIA H100 GPUs, and the runtime measurements in Sec.~\ref{sec:efficiency} use a single A100 GPU.

%% file: sec/B_streaming.tex
\section{Streaming QA Benchmarks}
\label{app:streaming}

\paragraph{Memoryless baseline.}
To better understand the behavior of existing streaming-QA benchmarks, we additionally evaluate a memoryless baseline that observes only the most recent frames. The prompt, input format, and timestamp representation are kept identical to those used by the other methods; the amount of temporal context available to the model is restricted. Following the setting of~\cite{simplestream}, we use a per-frame visual-token cap of 768. The recency-4f and recency-8f variants use 8 and 16 input frames at 2 fps, covering approximately four and eight seconds, respectively. The resulting KV caches require no eviction: their median lengths are 1{,}552 and 2{,}911 tokens on StreamingBench, or 19\% and 36\% of our budget of $M=8{,}192$, and 2{,}502 and 4{,}812 tokens on OVO-Bench, or 31\% and 59\%. Under this setting, we obtain 68.6 on OVO-Bench using the scoring convention of~\cite{simplestream}, closely matching their reported score of 67.7.

\input{tables/tableB_body}

\paragraph{Short-context baselines can outperform full memory.}
Table~\ref{tab:streaming} shows a markedly different behavior from the long-video benchmarks considered in the main evaluation. Short-context baselines achieve the highest aggregate scores across all three columns despite retaining only a small fraction of the available context. They also outperform Full KV, which retains the complete video history. This indicates that high aggregate scores on these benchmarks do not necessarily imply effective long-term memory retention.

The same pattern appears at the task level. Across the nine OVO-Bench tracks, recency-4f outperforms Full KV on seven, including optical character recognition (OCR, $+18.8$), object recognition (OJR, $+15.2$), action recognition (ACR, $+13.8$), and spatial understanding (STU, $+11.2$). These tracks primarily ask about information visible near the end of the stream, making recent observations sufficient for answering the question. In contrast, episodic memory (EPM, $-2.7$) and action sequence identification (ASI, $-8.8$) favor Full KV and require information from earlier portions of the video. Five of the ten StreamingBench tasks exhibit a similar pattern.

\paragraph{More context can degrade performance.}
Hallucination detection (HLD), which asks questions unrelated to the video and expects ``Unable to answer,'' provides the clearest example. Every method that processes the complete video scores below the random-guess expectation of 27.6\%, spanning 17.7 to 25.3: Full KV scores 18.3, ReKV 24.2, and DeltaS 19.9, whereas recency-4f reaches 43.0. Thus, low performance on this track cannot be interpreted solely as a failure to retain relevant history; access to substantially more context can reduce accuracy.

\paragraph{Performance on memory-dependent tasks.}
The relative ordering changes on tasks that require information from earlier in the stream. The Backward column of Table~\ref{tab:streaming} includes HLD. When HLD is excluded, DeltaS obtains the highest Backward score among all methods in the table, reaching 66.7 at $M = 8{,}192$, compared with 64.5 for Full KV, 59.8 for recency-4f, and 65.6 for InfiniPot-V, the next-best bounded-memory method at the same budget. At this budget, DeltaS also ranks first on EPM. On Counting, the StreamingBench task that requires accumulating information throughout the video, DeltaS reaches 60.1, whereas recency-4f and recency-8f score 39.4 and 39.9, respectively. Increasing the short-term observation window therefore provides little improvement, while retaining information across the stream yields a substantially higher score.

Overall, Table~\ref{tab:streaming} suggests that aggregate performance on StreamingBench and OVO-Bench is strongly influenced by tasks that can be solved from recent observations and, in some cases, become harder with longer context. We therefore report these benchmarks separately and use the memory-dependent tasks to examine whether an eviction method preserves information that must be revisited later.

%% file: tables/tableB_body.tex
\begin{table}[t]

\renewcommand{\band}[1]{\multicolumn{4}{l}{\textbf{#1}}\\[1pt]}%
\renewcommand{\hd}[1]{\makebox[48pt]{\makecell{#1}}}%
\centering
\footnotesize
\setlength{\tabcolsep}{3pt}
\begin{tabular}{@{\hspace{4pt}}l @{\hspace{4.4pt}} ccc@{\hspace{4pt}}}
\toprule
Method & \hd{Streaming\\Bench} & \hd{OVO\\Real-Time} & \hd{OVO\\Backward} \\
\midrule
\band{Recent frames only, no memory$^{\S}$}
\m{recency-4f} & 81.2 & 81.8 & 54.8 \\
\m{recency-8f} & 82.4 & 80.4 & 53.7 \\
\midrule
\band{Unbounded or query-aware}
\m{Full KV$^{\dag}$}  & 80.1 & 69.8 & 50.9 \\
\m{ReKV$^{\ddag}$}    & 77.0/78.2 & 66.4/69.2 & 52.8/52.6 \\
\midrule
\band{Simple baselines}
\m{random}   & 80.6/80.4 & 71.7/\underline{71.2} & 52.1/\textbf{53.1} \\
\m{recency}  & 80.4/\underline{80.5} & 72.0/70.0 & 50.7/51.5 \\
\m{uniform}  & 79.5/79.8 & 70.0/71.0 & 50.9/52.5 \\
\band{Streaming KV compression}
\m{InfiniPot-V} & \textbf{81.2}/\textbf{80.8} & \textbf{74.2}/\textbf{72.6} & \underline{52.8}/\underline{52.6} \\
\m{StreamMem}   & 77.0/76.4 & 64.5/64.9 & 51.0/50.2 \\
\m{HERMES}      & \underline{80.8}/80.0 & \underline{72.3}/71.1 & 52.1/52.3 \\
\rowcolor{oursbg}[4pt][4pt]
\m{\textbf{DeltaS (ours)}} & 80.2/79.9 & 71.3/70.1 & \textbf{52.9}/52.3 \\
\bottomrule
\end{tabular}
\caption{\textbf{Streaming QA benchmarks.}\quad Cells are $M = 8{,}192\,/\,16{,}384$. OVO-Bench is shown as its two multiple-choice regimes, each question-weighted. $^{\dag}$Full KV is unbounded. $^{\ddag}$ReKV uses the question to retrieve KV entries. $^{\S}$recency-4f/8f are not bound by $M$; their cache size is set by the frame window ($\approx$1.6k and 2.9k tokens on StreamingBench, 2.5k and 4.8k on OVO-Bench). Recent-frame-only, unbounded-memory, and query-aware reference methods are excluded from ranking.}
\label{tab:streaming}
\end{table}